\documentclass[sigconf,authorversion,nonacm]{aamas}

\usepackage{balance} 
\usepackage{subfigure}

\acmSubmissionID{890}

\title[Effectiveness of Potential-Based Reward Shaping]{Improving the Effectiveness of Potential-Based Reward Shaping in Reinforcement Learning}

\author{Henrik M\"uller}
\affiliation{
  \institution{L3S Research Center}
  \city{Hannover}
  \country{Germany}}
\email{hmueller@l3s.de}

\author{Daniel Kudenko}
\affiliation{
  \institution{L3S Research Center}
  \city{Hannover}
  \country{Germany}}
\email{kudenko@l3s.de}

\begin{abstract}
Potential-based reward shaping is commonly used to incorporate prior knowledge of how to solve the task into reinforcement learning because it can formally guarantee policy invariance. As such, the optimal policy and the ordering of policies by their returns are not altered by potential-based reward shaping.
In this work, we highlight the dependence of effective potential-based reward shaping on the initial Q-values and external rewards, which determine the agent's ability to exploit the shaping rewards to guide its exploration and achieve increased sample efficiency.
We formally derive how a simple linear shift of the potential function can be used to improve the effectiveness of reward shaping without changing the encoded preferences in the potential function, and without having to adjust the initial Q-values, which can be challenging and undesirable in deep reinforcement learning. 
We show the theoretical limitations of continuous potential functions for correctly assigning positive and negative reward shaping values. 
We verify our theoretical findings empirically on Gridworld domains with sparse and uninformative reward functions, as well as on the Cart Pole and Mountain Car environments, where we demonstrate the application of our results in deep reinforcement learning.
\end{abstract}


\newcommand{\BibTeX}{\rm B\kern-.05em{\sc i\kern-.025em b}\kern-.08em\TeX}

\begin{document}


\pagestyle{fancy}
\fancyhead{}


\maketitle 


\section{Introduction}
Reward shaping is a common approach to accelerate the training of reinforcement learning agents by incorporating some form of external guidance into the reward function, thereby improving the exploration of the environment.

In this work, we focus on potential-based reward shaping \citep{ng1999invariance}. The primary appeal of potential-based reward shaping is the guarantee of policy invariance. Despite changing the rewards, the optimal policy of the MDP with the shaped reward function remains identical to that of the original MDP. Potential-based reward shaping utilizes a potential function to assign a heuristic value of \textit{goodness} (or potential) to each state, with the reward shaping subsequently derived from the difference between the potential of the states before and after executing an action.

Previous theoretical evaluations of potential-based reward shaping have given pointers of how to structure an effective potential function \citep{grzes2009potential-function-analysis, grzes2017episodicPBRS}, yet they have overlooked the intrinsic link between the reward and the initialization of the state-action values. 
While potential-based reward shaping has been shown to be equivalent to shifting the Q-value initialization by adding the potential function \citep{wiewiora2011Qinitialization}, previous work has not addressed two key questions: first, how the Q-value initialization and external rewards affect sample efficiency in potential-based reward shaping; and second, how the potential function can be modified to improve the sample efficiency without altering the encoded preferences over states.

In this paper, we propose a modification of the potential function incorporating a constant bias $b$. 
We demonstrate that a constant bias of the potential function can be used in sparse-reward settings to improve the sample efficiency by improving the attribution of positive and negative shaped rewards through the mitigation of differences between the intermediate rewards and the initial Q-values.
We show that scaling the potential function is inherently limited in its ability to create a more effective reward shaping in terminal MDPs.

Notably, potential-based reward shaping does not alter the optimal policy, and our method does not change the state preferences of the potential function. Consequently, our approach is applicable in any situation where additional knowledge of the task can be exploited for more sample-efficient RL.

The primary contributions of this paper can be summarized as follows:
\begin{enumerate}
    \item We introduce a generalized framework of requirements for an effective potential-based reward shaping.
    \item We explore how to choose the scale and offset for potential functions to adapt the potential function to the Q-value initialization and reward function, thus improving the effectiveness of reward shaping.
    \item We show the theoretical limitations for the creation of effective potential-based reward shaping when dealing with continuous potential functions.
    \item We verify our findings empirically first on tabular RL, and then extend our experiments to the deep RL setting.
\end{enumerate}


\section{Related Work}
Potential-based reward shaping (PBRS) is a type of reward shaping defined by the difference in heuristic valuations of the states before and after the execution of an action using a potential function over the state space. PBRS is widely used, because it has been shown to be necessary and sufficient for policy invariance \citep{ng1999invariance}. 
In \citet{wiewiora2011Qinitialization}, it was shown that potential-based reward shaping is equivalent to simply adding the potential function values to the corresponding initial Q-values and then continuing the learning process with the original non-shaped reward function. However, in the case of potential functions that change dynamically over time, it was shown that while policy invariance still holds, the change in the learning process cannot be equated to a simple modification of the initial Q-values \citep{devlin2012dynamicPBRS}.
This work extends the results of \citet{grzes2009potential-function-analysis} on considerations for effective PBRS to include the connection between reward shaping and the initial Q-values and external rewards.

PBRS is often used to incorporate prior knowledge of the task to guide the agent in solving difficult tasks. The potential function can be created from automatically extracted automata specifying a sequence of goals to achieve \citep{Hasanbeig2021DeepSynthAS}, from user provided linear temporal logic formulas \citep{elbarbari2022tlrl}, or directly from demonstrations in imitation learning settings \citep{wang2023dshape, brys2015demonstration, suay2016demonstration, wu2021demonstrations}.

In \citet{sun2022optimistic}, the authors show that a constant linear reward shift is equivalent to an optimistic or pessimistic Q-value initialization, depending on the value of the reward shift. Our work can provide a novel perspective on this problem. The constant reward shift by a constant $c$ is equivalent to using a constant potential function $\Phi(s)= \dfrac{c}{\gamma - 1}$ in PBRS. By linking to potential-based reward shaping, we can directly draw on its large catalog of theoretical results regarding policy invariance \citep{ng1999invariance}, the connection to Q-value initialization \citep{wiewiora2011Qinitialization}, and the need for potential values of zero in terminal states \citep{grzes2017episodicPBRS}. Notably, the last point is missing in \citet{sun2022optimistic}. 
Therefore, using the same constant shift for all rewards can change the optimal policy when moving into terminal states of episodic MDPs. We are able to extend their results to also include additional knowledge about how to solve the task for more fine-grained guidance about which parts of the state space to explore and which parts to avoid exploring.


\section{Background}
\subsection{Reinforcement Learning}
\label{sec:rl}
Reinforcement learning (RL) leverages experience from interactions with its environment to solve sequential decision-making tasks. At each step, an RL agent can execute an action based on the current environment state and will receive the reward for the executed action and the new environment state.
Formally, the problem for RL is commonly defined as a Markov decision process (MDP). An MDP is a tuple $M = (S, A, T, R, \gamma)$, where $S$ is the state space, $A$ is the action space, $T$ describes the transition dynamics when executing action $a$ in state $s$, $R$ is the reward function attributing scalar reward feedback to transitions, and $\gamma$ is the discount factor used to define the goal to maximize - the discounted return $\sum_t \gamma^t r_t$.

We focus in this work on sparse reward functions that offer very little (intermediate) feedback for the exploration strategy of the learning agent and are therefore inherently difficult to solve efficiently. We adopt the reward formulation of \citet{matignon2006RewardFA} for a goal-directed reward function with respect to a goal state $s_g$: 
\begin{equation} \label{eq:goal-directed-reward}
R(s,a,s') = 
    \begin{cases}
        r_g &\text{if } s' = s_g \\
        r_\infty &\text{otherwise}
    \end{cases}
\end{equation}

In section \ref{experiments}, we will also empirically verify our theoretical findings on two different well-known parameterizations of this reward function. The goal-directed reward where $r_g =1$ and $r_\infty = 0$ and the on-step reward where $r_g = r_\infty$.
In goal-directed tasks, the only non-zero reward is given for reaching a goal state, where the episode terminates. Tasks with uniform on-step reward functions always give the same non-zero reward for each transition. Their goal is either to maximize the number of steps before termination for positive reward values, or to minimize the number of steps before reaching a terminal state for negative rewards.


\subsection{Potential-Based Reward Shaping}
Potential-based reward shaping (PBRS) is defined by its potential function $\Phi(s)$ mapping each state to a heuristic scalar value. Given the potential function, the reward shaping function $F$ is defined as:
\begin{equation}
F(s, a, s') = \gamma \Phi(s') - \Phi(s)
\end{equation}
where $s'$ is the state reached after executing action $a$ in $s$ and $\gamma$ is the reward discount factor of the MDP. With this the shaped reward $R'$ can be defined as:
\begin{equation}
R'(s, a, s') = R(s, a, s') + F(s, a, s')
\end{equation}

Due to the guarantee of policy invariance for potential-based reward shaping, the optimal policy (and the order of returns for all policies) will not change for the shaped MDP $M' = (S, A, T, R', \gamma)$ \citep{ng1999invariance}. As such, there is no way to specify a reward shaping function without making assumptions about the reward and transition functions that does not alter the optimal policy except through shaping functions that are equivalent to PBRS. As a result, PBRS can guarantee not to change the problem defined by the reward function and can instead be used to add exploitable structure to the reward function of an MDP to guide the exploration.

In the case of episodic MDPs, it is necessary that all Q-values of terminal states be equal to zero. As a consequence of the connection between PBRS and the Q-value initialization, the potential function must also be equal to zero in all terminal states to be able to guarantee policy invariance \citep{grzes2017episodicPBRS}. Notably, this includes the state preceding the truncation of a training episode after a fixed number of steps.
Intuitively, potential-based reward shaping removes the previously added potential by subtracting it in the very next transition to remove the reward shaping from the total return. If the episode ends, there is no next transition and the potential of the last (terminal) state in the episode has to be zero to avoid biasing the total return, which could change the optimal policy. Nonetheless, all of our conclusions will extend to non-episodic MDPs unless they explicitly focus on the special case of terminal states.

Although, PBRS can guarantee not to change the optimal policy (and - if applicable - the convergence guarantee to the optimal policy of the used RL agent), the convergence speed and therefore the sample efficiency can vary depending on how the potential function is defined. 
In this work, we focus on how to structure potential functions to create the most effective bias of a learning agent rather than on how to create and ground potential functions.


\section{Motivating Example}
\label{sec:motivation}

\begin{figure}[ht]
\centering
\includegraphics[width=\columnwidth]{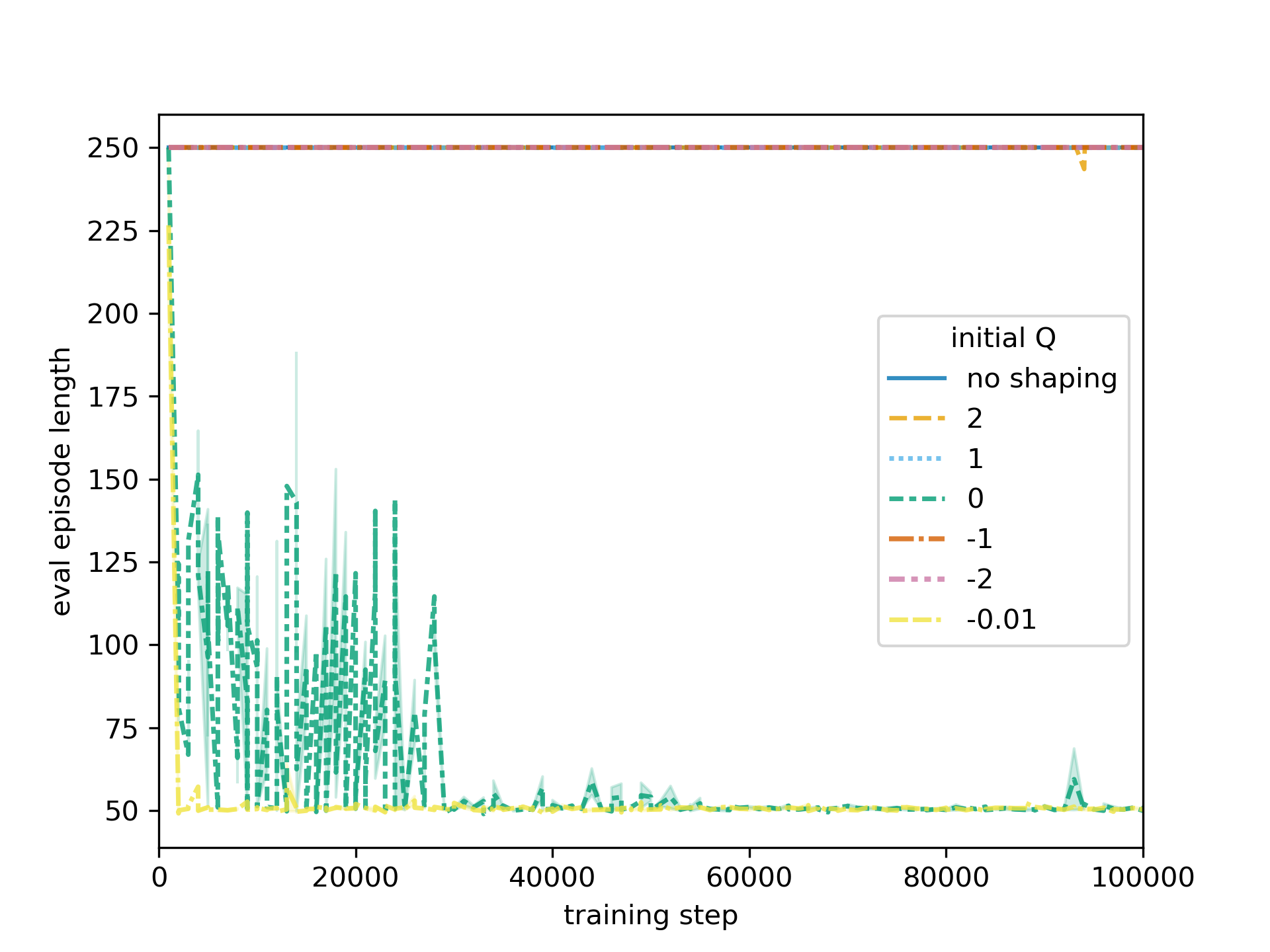}
\caption{Average length of evaluation runs (with $\epsilon=0.05$) on a 25x25 Gridworld with potential-based reward shaping where $\Phi(s) = V^*(s)$.}
\label{fig:vstar-gridworld-results}
\Description[Average length of evaluation runs with respect to the initial Q-values on a Gridworld for potential-based reward shaping where $\Phi(s) = V^*(s)$.]{Average length of evaluation runs (with $\epsilon=0.05$) on a 25x25 Gridworld with potential-based reward shaping where $\Phi(s) = V^*(s)$ for different initial Q-Values. The no-shaping baseline and initial Q-values $2$, $1$, $-1$, and $-2$ do not converge during the entire training of 100,000 steps. Zero initialization leads to slow, unstable convergence over the first 25,000 steps. Initializing with $-0.01$ leads to very quick convergence to the optimal solution.}
\end{figure}

We first look at the dependence of effective potential-based reward shaping on the initial Q-values when using the optimal state value function $V^*$ as the potential function as introduced in \citet{ng1999invariance}.
Intuitively $V^*$ should be the best possible state-based guidance one can offer.
If one uses $\Phi(s) = V^*(s)$, the shaped rewards will be exactly zero for any optimal action and less than zero for any sub-optimal action. The RL agent therefore should only have to explore to determine which actions give a reward of zero and should then be able to exploit this knowledge to bias the action selection to repeat these actions.

But whether the RL agent is able to exploit to shaped rewards to easily recognize optimal actions relies entirely on the Q-Value initialization, which (although generally well established) has not been considered in the context of potential-based reward shaping. If all Q-Values are larger than zero, the agent would have to initially try every action (possibly multiple times) to correctly discern the optimal action. If the Q-Values are all smaller than zero the agent could start to focus on whichever sub-optimal action were randomly explored early on. In that case, the agent would depend on exploration actions whilst the exploitative (greedy) action selection would select sub-optimal actions. 

For the fastest convergence the optimal initialization would therefore be with a value between zero (for the shaped rewards of all optimal actions) and the smallest possible shaping difference with respect to the given $V^*$. As soon as an optimal action is first encountered the Q-Value of that action would increase and end up being the largest Q-Value for that state. Any of the commonly used advantage-based action selection schemes (like $\epsilon$-greedy) would then always choose the optimal action in that state (outside of the random action selections for exploration).

In figure \ref{fig:vstar-gridworld-results} we show the average episode length for ten evaluation runs every 500 training steps in a simple 25-by-25 Gridworld. The task of the agent is to move from the top left into the bottom right corner of the grid. The agent gets a reward of zero for every action except when moving into the goal. Initializing the Q-Values with $-1$ or $1$ leads to the gray and orange lines respectively that do not converge within the 100,000 training steps. The blue graph shows the behavior for a zero initialization. It converges fast, but takes many more steps to stabilize in the evaluation performance as unexplored actions in states that are reached by executing a small number of random steps during exploration cannot be differentiated from already found optimal actions. The green plot for an initialization of all Q-Values with $0.01$ solves this problem by being pessimistic with regard to the optimal actions. 

Overall, this highlights the importance of the initialization for the convergence. In the following we will explore how to adapt a given potential function to gain the same benefits as a good initialization without having to adapt the initialization itself, which is not always trivial. Especially, when using a function approximator as in deep RL.


\section{Main Results}
\label{sec:effective-bias}
Potential-based rewards shaping can be used to increase the intermediate valuation of specific actions and transitions at the beginning of the learning process while the guaranteed policy invariance will ensure the eventual convergence towards the original valuation of policies. 
The goal for an effective application of PBRS is for the agent to exploit the potential function to guide its initial exploration, thereby opting to maximize the value of the potential function at the start of the training. As such, our theoretical results will focus on the first updates of the initial estimates.

To obtain concrete results, we exemplarily focus here on the update in Q-Learning (similar results can be obtained analogously for other TD-based approaches):
\[
Q(s_t, a_t) \leftarrow Q(s_t, a_t) + \alpha \left[ R_{t+1} + \gamma \max_a Q(s_{t+1}, a) - Q(s_t, a_t) \right]
\]
The target for the TD-update $R_{t+1} + \gamma \max_a Q(s_{t+1}, a)$ determines whether the respective Q-value is increased or decreased from the initial value.
In advantage-based action selection, this results in the action being more likely to be repeated or respectively ignored when the same state is visited again. 

Notably, for the first updates on the initial Q-values we have:
\[
Q(s_t, a_t) = \max_a Q(s_{t+1}, a) = Q_{init}
\]

Therefore, we can simplify the TD-update and focus on the following case instead: $R'(s, a, s') + \gamma Q_{init}$. 
We can now formalize the requirements for increasing and decreasing Q-values with the following equation:
\begin{eqnarray}
    R'(s, a, s') + \gamma Q_{init} \lessgtr Q_{init} \\
    \therefore R'(s, a, s') \lessgtr (1 - \gamma) Q_{init}
\end{eqnarray}

The general idea is that the shaped reward should be larger (or respectively smaller) than the initial Q-value for transitions that should be incentivized (or respectively disincentivized) in the future. After the transition is seen for the first time the updated Q-value would then be larger (or smaller) than the initial Q-value and any advantage-based action selection mechanism (like $\epsilon$-greedy) would then prefer (or avoid) the corresponding action. That way excessive exploration can be avoided and the agent focuses in the initial exploration on states with increasingly larger potential values.

The following set of requirements for the relation of the shaped rewards to the initial Q-values is an extension of the requirements originally proposed in \citet{grzes2009potential-function-analysis} using the above relation between TD-update and initial Q-values, where we look at transitions between the states $s$ and $s'$ with $\Phi(s') > \Phi(s)$:
\begin{eqnarray}
\label{eq:pot-req1}
r_\infty + \gamma \Phi(s') - \Phi(s) > (1 - \gamma) Q_{init} \\
\label{eq:pot-req2}
r_\infty + \gamma \Phi(s) - \Phi(s') < (1 - \gamma) Q_{init} \\
\label{eq:pot-req3}
r_\infty + \gamma \Phi(s) - \Phi(s) \leq (1 - \gamma) Q_{init}
\end{eqnarray}

These three inequalities formalize the requirements on non-terminal transitions. Only transitions that lead to states with a higher potential value should be incentivized. Other transitions should be disincentivized at the beginning of the learning process. 
As a result, any of the commonly used advantage-based action selection schemes would repeat actions that lead to the largest possible next potential value while avoiding to repeatedly explore actions that lead to states with smaller potential values.

\subsection{Potential Scale}
For goal-directed MDPs, we have the following additional requirements on the value of the shaped rewards when moving from any state $s$ into a terminal state:
\begin{eqnarray}
\label{eq:pot-req4}
r_g - \Phi(s) > (1 - \gamma) Q_{init} \\
\label{eq:pot-req5}
r_\infty - \Phi(s) < (1 - \gamma) Q_{init}
\end{eqnarray}

Intuitively, reaching the goal state should be incentivized. Reaching any other terminal state or truncating the episode before reaching the goal state should be disincentivized. As a direct result of these two inequalities, we can obtain the upper and lower bounds on the scale of the potential function in goal-directed MDPs:
\begin{equation}
\label{eq:potbounds}
r_\infty - (1 - \gamma) Q_{init} < \Phi(s) < r_g - (1 - \gamma) Q_{init}
\end{equation}

As a direct result of these bounds, it is not possible to utilize the scale of the potential function to compensate the mismatch between the original rewards and the initial Q-values, and thus satisfying the general requirements outlined in equations \ref{eq:pot-req1}-\ref{eq:pot-req3}.

Notably, equation \ref{eq:potbounds} implicitly requires that $r_g > r_\infty$. Accordingly, the incentive to terminate in a goal state must originate from the original reward of a goal-directed MDP, as potential-based reward shaping cannot create this incentive.

The goal-directed reward specification detailed in section \ref{sec:rl} therefore only allows for a difference of one between the smallest and largest potential values. The on-step reward does not offer a different reward for reaching a goal state. Consequently, it can result in the creation of incorrect shaped rewards when moving into a terminal state. In this case, depending on the difference between the reward and the Q-values, the agent therefore is either incentivized to focus on terminating in any (previously explored) terminal state or disincentivized from terminating in any terminal state, including possible goal states.

\subsection{Potential Shift}
\label{sec:shift}
The equations \ref{eq:pot-req1} to \ref{eq:pot-req3} show, that the effectiveness of PBRS depends not just on the potential function, but also on the external reward and the initial Q-values. If the initial Q-values and the (mostly constant) original rewards are known, one could add a simple bias to the potential function to remove the dependence. 
We define the shifted potential function for any non-terminal state $s$ as:
\begin{equation}
    \label{eq:pot-bias}
    \Phi_b(s) = \Phi(s) + \frac{b}{\gamma - 1}    
\end{equation}
This constant bias term shifts all rewards (except when moving into terminal states) by $b$. If we set $b = (1 - \gamma) Q_{init} - r_\infty$, we are able to remove the dependence on both the external reward and the initial Q-values. This allows us to make direct use of the prior results on how to create an effective PBRS \cite{grzes2009potential-function-analysis, mueller2025epbrs}.

This shift of the potential values has to exclude the potential values for terminal states as the potential of any terminating state has to be zero \cite{grzes2017episodicPBRS}. The above choice for the bias term therefore cannot remove the dependence on the external reward and Q-values in equation \ref{eq:pot-req4} and \ref{eq:pot-req5}. 
For these transitions the additional term in the shaped reward is $Q_{init} + \frac{r_\infty}{\gamma - 1}$.
The benefit of removing $r_\infty$ and $Q_{init}$ in the requirements for non-terminal transitions therefore can come at the cost of incorrectly (dis-)incentivizing transitions into terminal states.
This can create challenges during training where all terminal states either have a (large) positive or (large) negative shaping reward. Which causes the agent to focus its exploration on terminating in any possibly sub-optimal state or lead the exploration away from any terminal state. 
The application of our method is therefore more limited in settings with short episode lengths where episodes are more likely to be truncated and in settings where the difference between terminal states is important (e.g. when learning to reach a terminal goal state).


\begin{figure*}[ht]
\centering
\subfigure[linear potential]{\includegraphics[width=0.3\textwidth]{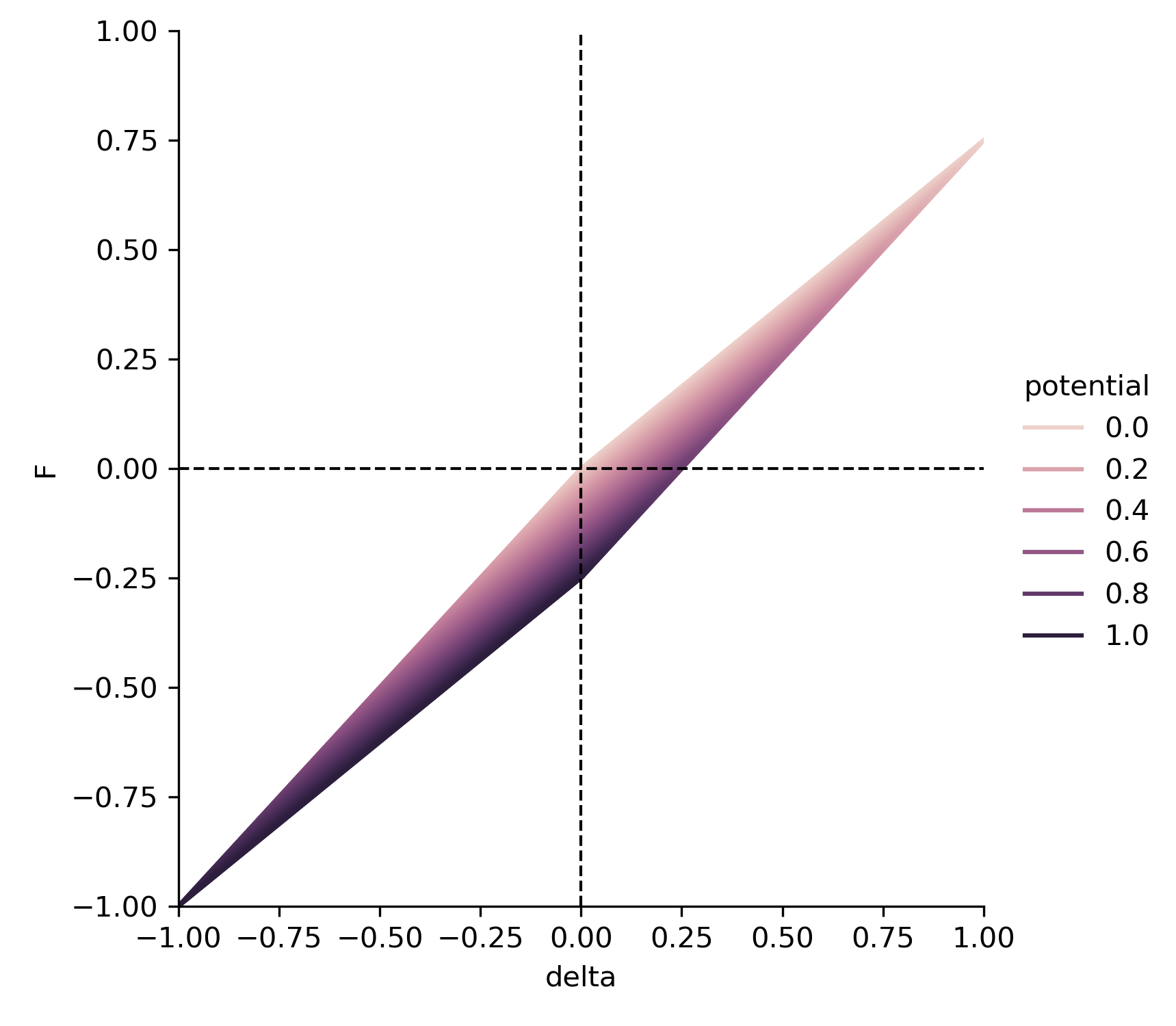}}
\subfigure[exponential potential with $e=8$]{\includegraphics[width=0.3\textwidth]{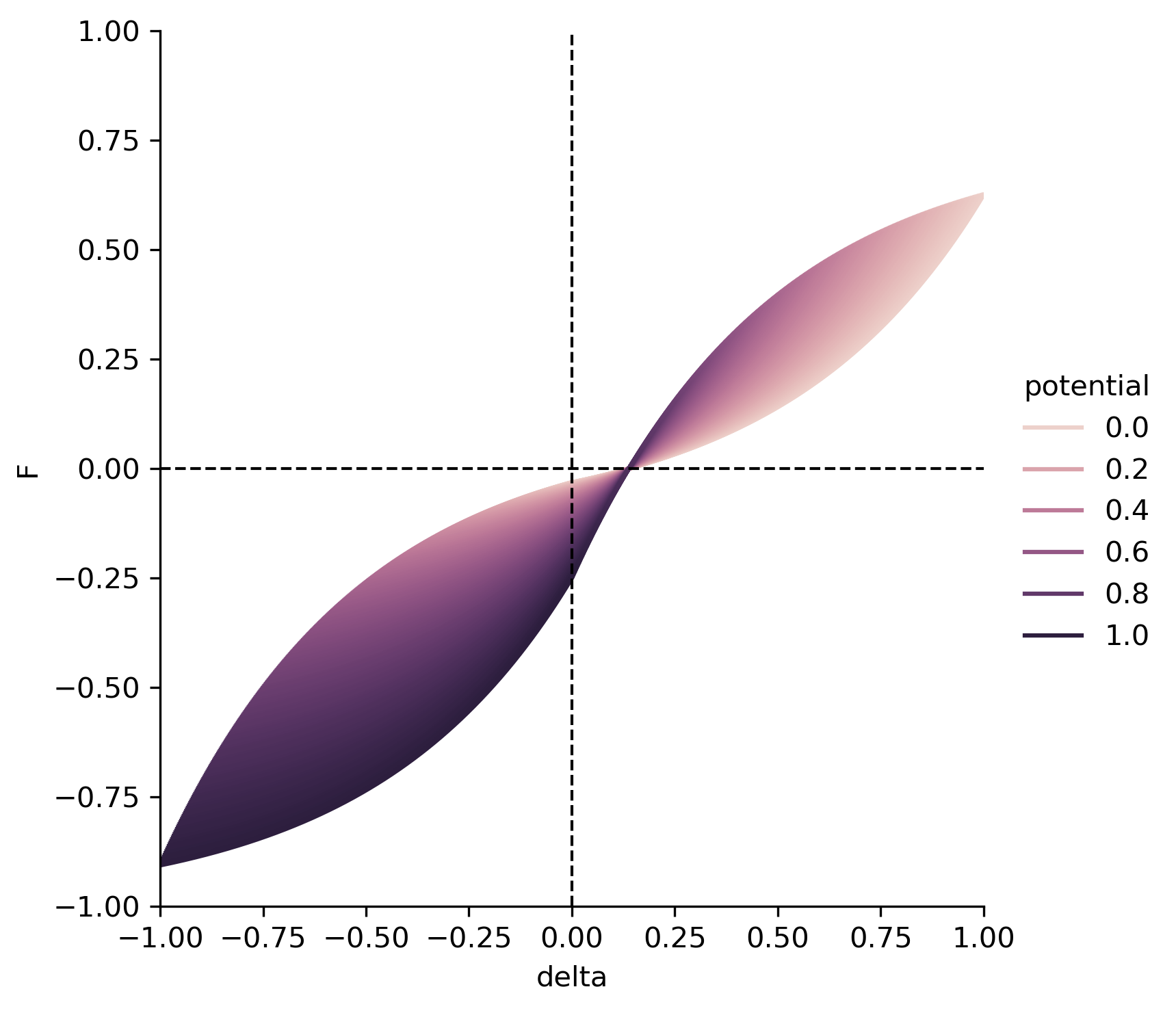}}
\subfigure[exponential potential with $e=64$]{\includegraphics[width=0.3\textwidth]{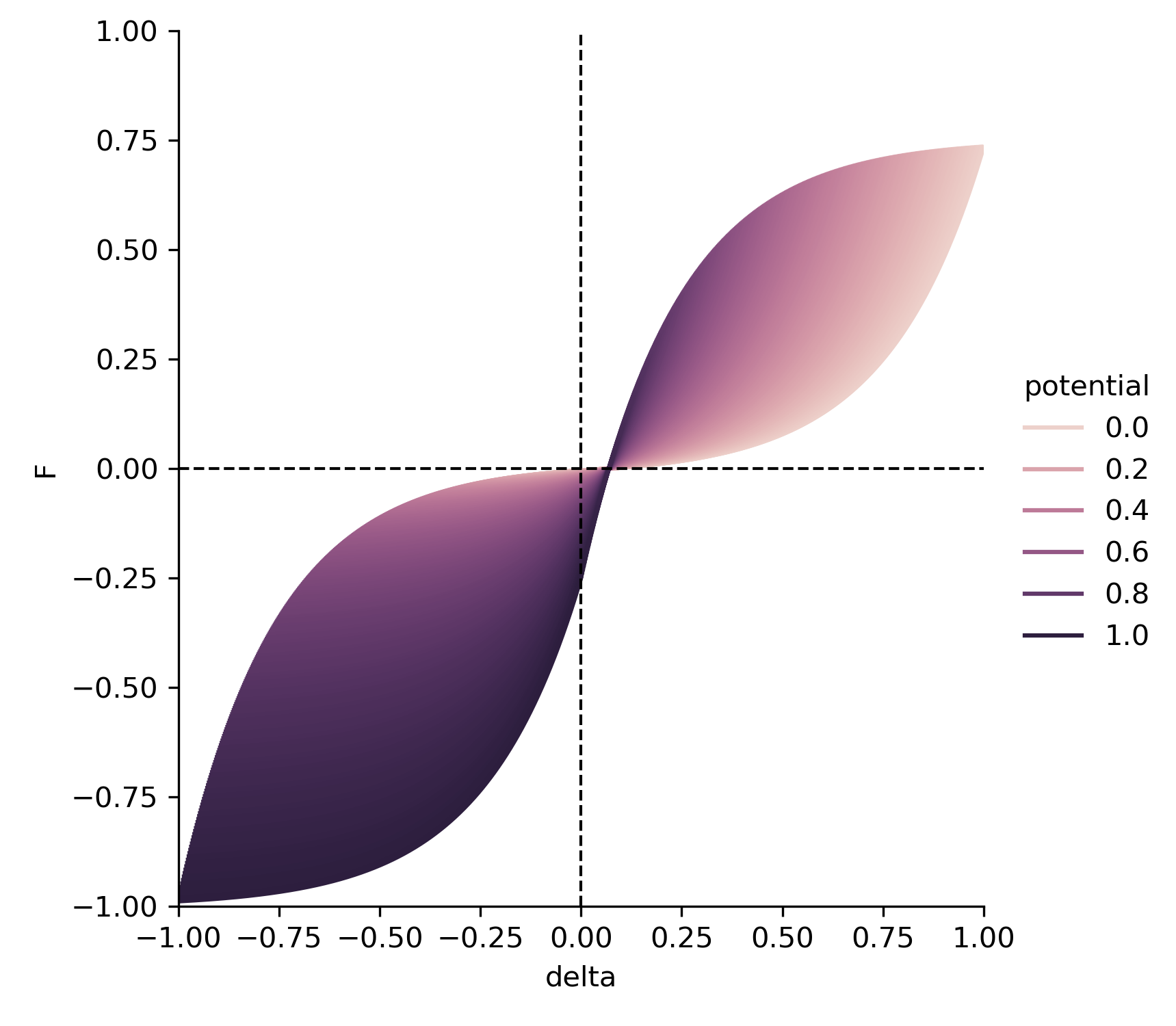}}
\caption{Plots of the reward shaping F that will be added to the reward given the difference $\delta$ between the next potential and the previous potential with the hue representing the value of the previous potential. All plots are for $\gamma = 0.75$. For the plots of exponential PBRS the difference $\delta$ is defined as the difference in the original (linear) potential functions.}
\label{fig:F-plots}
\Description[Shaping rewards F given differences in the potential and given the previous potential value for difference potential functions (linear, exponential).]{Plots of the reward shaping F that will be added to the reward given the difference $\delta$ between the next potential and the previous potential with the hue representing the value of the previous potential.}
\end{figure*}

\subsection{Limitations of continuous potential functions}
\label{sec:perfect-potential}

As shown in \citet{grzes2009potential-function-analysis}, any continuous potential function will create some incorrect shaping rewards. In positive potential functions this occurs in the form of negative shaping rewards for small improvements $\delta$ in potential between consecutive states:
\begin{equation}
\gamma \Phi(s') - \Phi(s) < 0\\
\end{equation}    
\begin{equation}
\gamma (\Phi(s) + \delta) - \Phi(s) < 0\\
\end{equation}
\begin{equation}
\delta < \frac{1-\gamma}{\gamma}\Phi(s)
\end{equation}

In negative potential functions this instead occurs in the form of positive shaping rewards for small decreases $\delta$ in potential between consecutive states:
\begin{equation}
\gamma \Phi(s') - \Phi(s) > 0\\
\end{equation}
\begin{equation}
\gamma (\Phi(s) - \delta) - \Phi(s) > 0\\
\end{equation}
\begin{equation}
\delta > \frac{\gamma - 1}{\gamma}\Phi(s)
\end{equation}

As a result, it is in general impossible for continuous potential functions to create both of the required shaping rewards correctly at the same time and there will be some difference $\delta$ for which the equations \ref{eq:pot-req1} and \ref{eq:pot-req2} will be violated. 
Additionally, the smallest value with correct positive or respectively negative shaping rewards depends on the potential of the state the transition started in. Small improvements in potential are therefore more likely to be create incorrect incentives in states with a smaller potential value (e.g. at the beginning of an episode) for positive potential functions and in states with potentials closer to zero (e.g. close to the goal) for negative potential functions.

In \citet{mueller2025epbrs}, it was shown that the dependence on the previous potential can be solved in a setting with a discrete potential function by employing an exponentially growing potential function instead to guarantee correct exploration incentives.

The advantage of exponentially growing potential functions is also shown in figure \ref{fig:F-plots}. It plots the shaping reward $F$ for the potential value difference $\delta$ between two consecutive states with respect to the potential value before the transition. Intuitively, the top-right and bottom-left quadrants are (with respect to the potential function) correctly incentivized transitions. Notably, larger values of the base of the exponent can support smaller potential differences $\delta$.

We therefore adapt the exponentially growing potential function and set $e^{\Phi(s')} = e^{\Phi(s) + \delta}$, where $\delta > 0$ is the change of the potential function when moving into the new state and where the base $e > 1$ is a hyper-parameter. This leads to the following set of requirements:
\begin{eqnarray}
r_\infty + (\gamma e^\delta - 1) e^{\Phi(s)} > (1 - \gamma) Q_{init} \\
r_\infty + (\gamma - e^\delta) e^{\Phi(s)} < (1 - \gamma) Q_{init} \\
r_\infty + (\gamma - 1) e^{\Phi(s)} \leq (1 - \gamma) Q_{init} 
\end{eqnarray}

Notably, if $(1-\gamma)Q_{init} - r_\infty = 0$ (or if we include a constant bias in the potential to match the difference) one can divide by the previous potential $e^{\Phi(s)}$ and therefore remove the dependency on the value of the previous potential from the first three inequalities. 
The correct value of the shaped reward is then only dependent on the value of the change $\delta$ between two consecutive states. In this case it is possible to determine a smallest change between the potential values of two consecutive states that will still correctly bias the intermediate Q-value to be more or respectively less likely to be picked in future action selections.

In practice, it would then intuitively also be possible to keep track of the smallest observed difference in potential between consecutive states and choose an appropriate base parameter $e$ dynamically. But changing the potential function while trying to still use the already (partially) converged Q-values would be a challenge for effective PBRS. As seen before, the effectiveness of PBRS heavily depends on the Q-values and continuing to learn with a changed potential function could lead to the Q-values being either too large or too small to make effective use of the potential function. Ideally you would not have to reset Q-values and should be able to make use of your prior experiences, though we leave the details of how to still guarantee effective convergence and obtain high sample efficiency for dynamically changing potential functions for future work.

If we have $(1-\gamma)Q_{init} - r_\infty \neq 0$ and the bias term is not chosen correctly, the smallest supported difference $\delta$ with correct (dis-)incentivizations of transitions would then also still depend on the potential of the state before the transition. But the smallest supported difference would scale with the logarithm of the potential, which would still be beneficial over the linear scaling in linearly growing potential functions.


\section{Experiments}
\label{experiments}

\subsection{Algorithms and Hyperparameters}
\subsubsection{Q-Learning}
For the experiments on the Gridworld environment we use a simple tabular Q-Learning agent. This allows to verify our theory in a simple setting on a well established model-free RL algorithm without the additional complexity of function approximation. We use a fixed exploration rate of $\epsilon = 0.05$ and a learning rate of $0.1$.

\subsubsection{DQN}
In the experiments on the Cart Pole and Mountain Car environments we use DQN \citep{mnih2013dqn} as a representative algorithm for deep RL methods. We utilize DQN as a proxy for Q-Learning in deep RL, thereby circumventing the potential confounding effects of numerous additional design parameters. We use the implementation of Stable Baselines3 \citep{stable-baselines3} for DQN including the default set of hyperparameters for the algorithm.

The exploration rate $\epsilon$ for DQN decays over the first 10,000 steps from $1$ to $0.05$. The learning starts after the first 1000 training steps have been collected with a learning rate of $0.0001$ and a batch size of $32$. The maximum replay buffer size is set to 50,000. The experiments on the Cart Pole and Mountain Car environments with the DQN use a decay factor of $\gamma = 0.99$.

\subsubsection{Potential Function Hyperparameter} For the reasons described in section \ref{sec:perfect-potential}, we will use the exponentially growing potential functions for all experiments with the base parameter set to $e=32$. Larger values for $e$ did not show any different results in our experiments.

\subsection{Environments and Results}
\begin{figure*}[htb]
\centering
\subfigure[goal-directed, $Q_{init}$ = -1]{\includegraphics[width=0.33\textwidth]{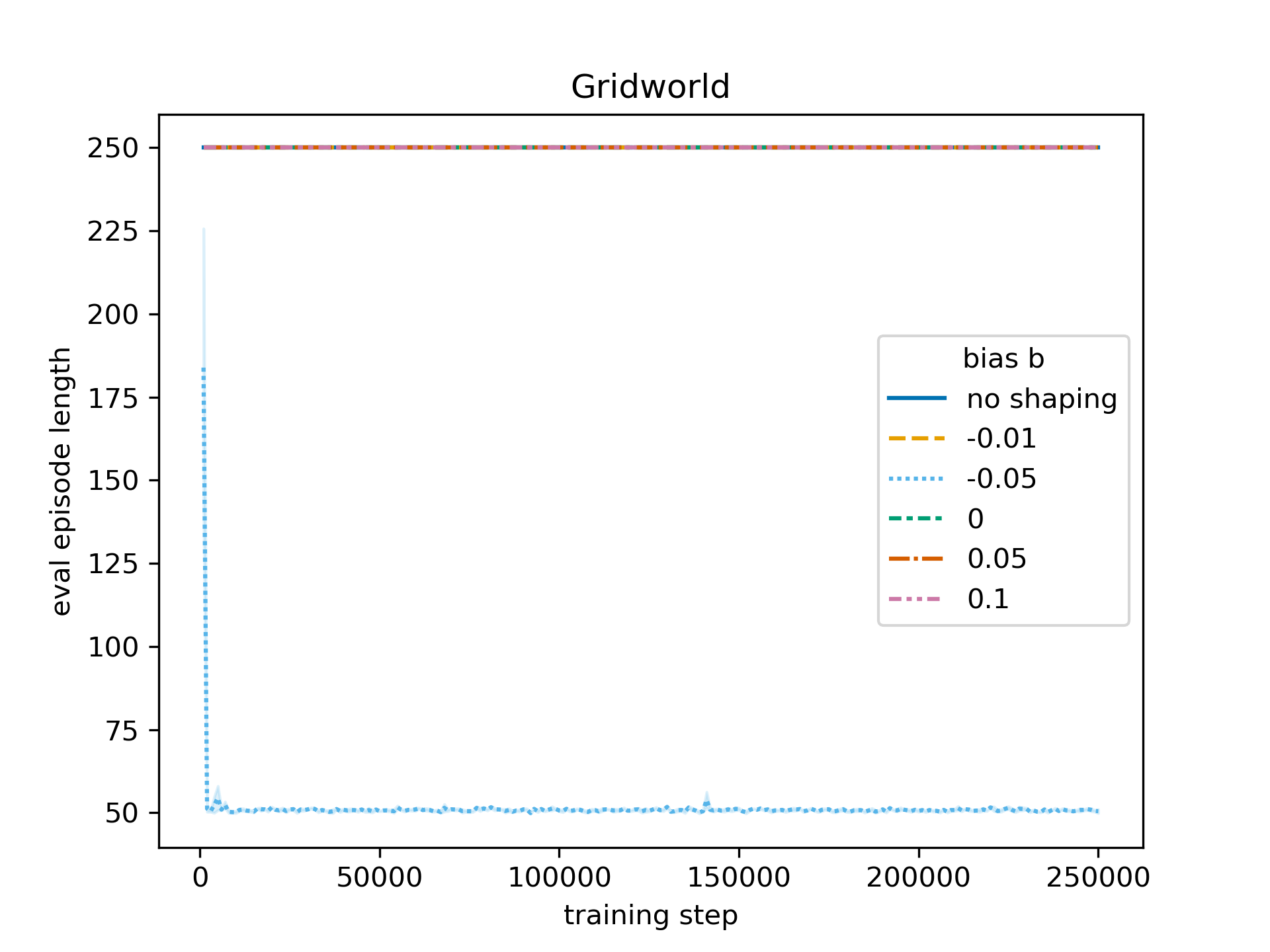}}
\subfigure[goal-directed, $Q_{init}$ = 0]{\includegraphics[width=0.33\textwidth]{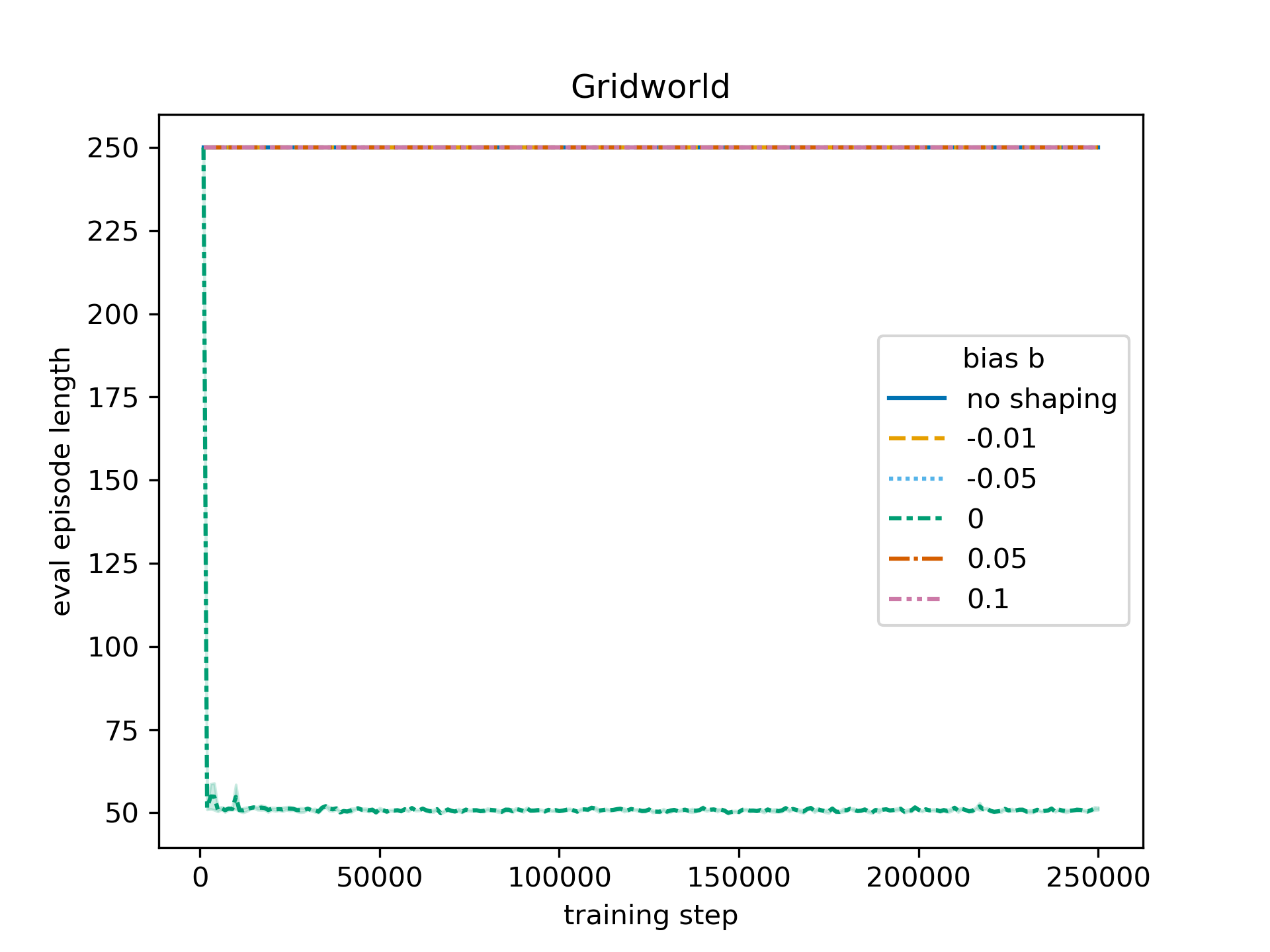}}
\subfigure[goal-directed, $Q_{init}$ = +1]{\includegraphics[width=0.33\textwidth]{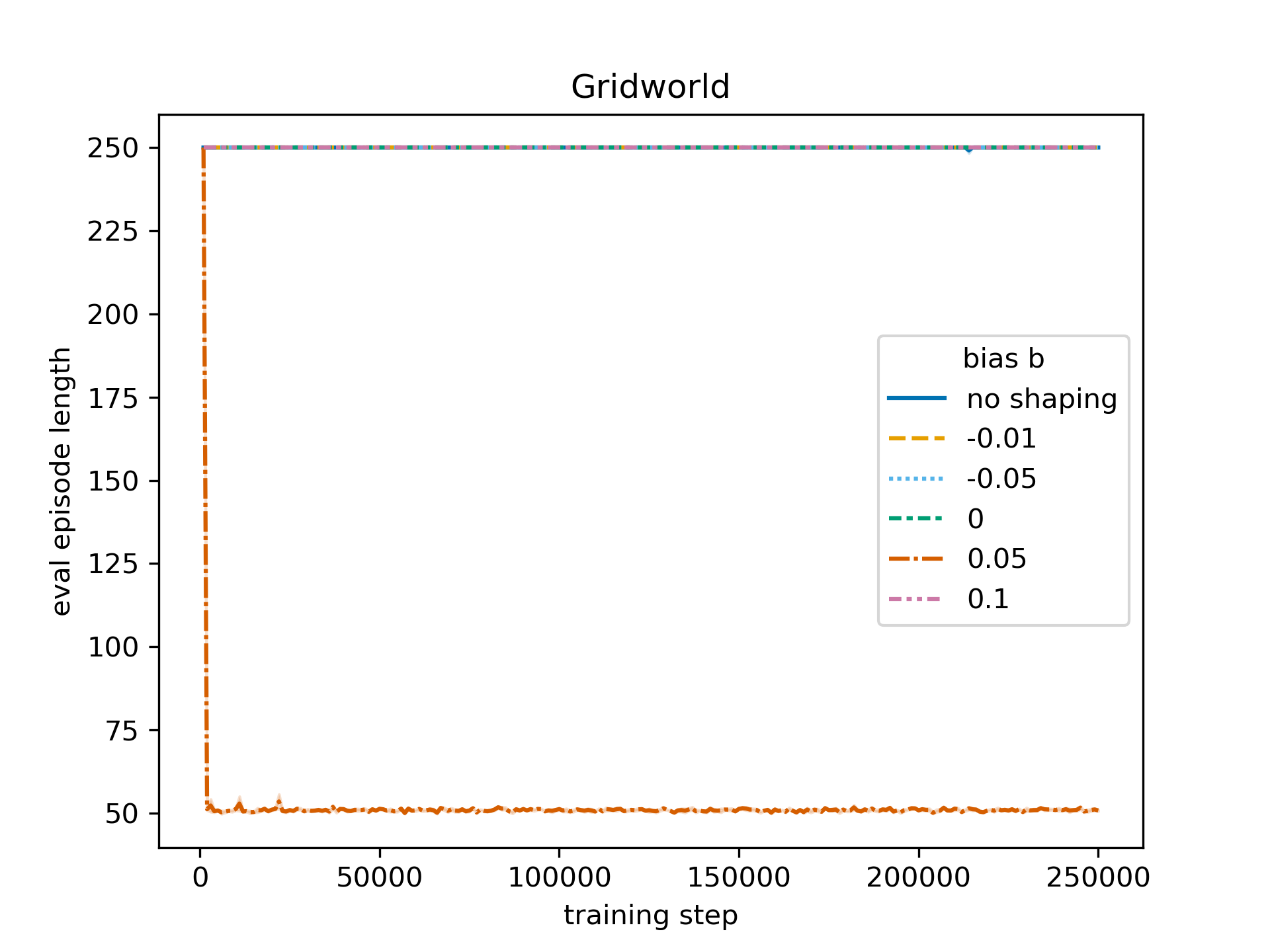}}
\\[\smallskipamount]
\subfigure[on-step, $Q_{init}$ = -1]{\includegraphics[width=0.33\textwidth]{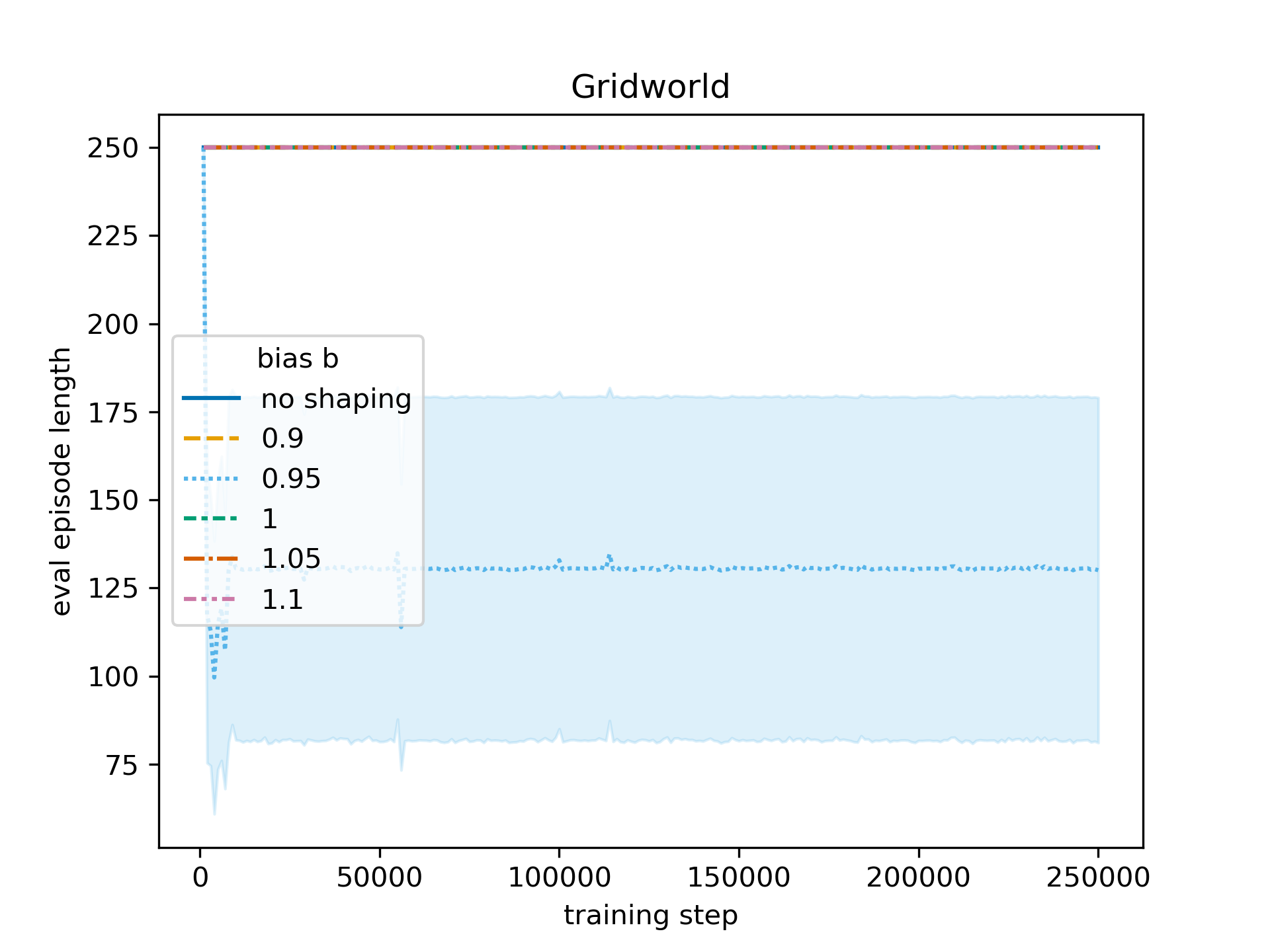}}
\subfigure[on-step, $Q_{init}$ = 0]{\includegraphics[width=0.33\textwidth]{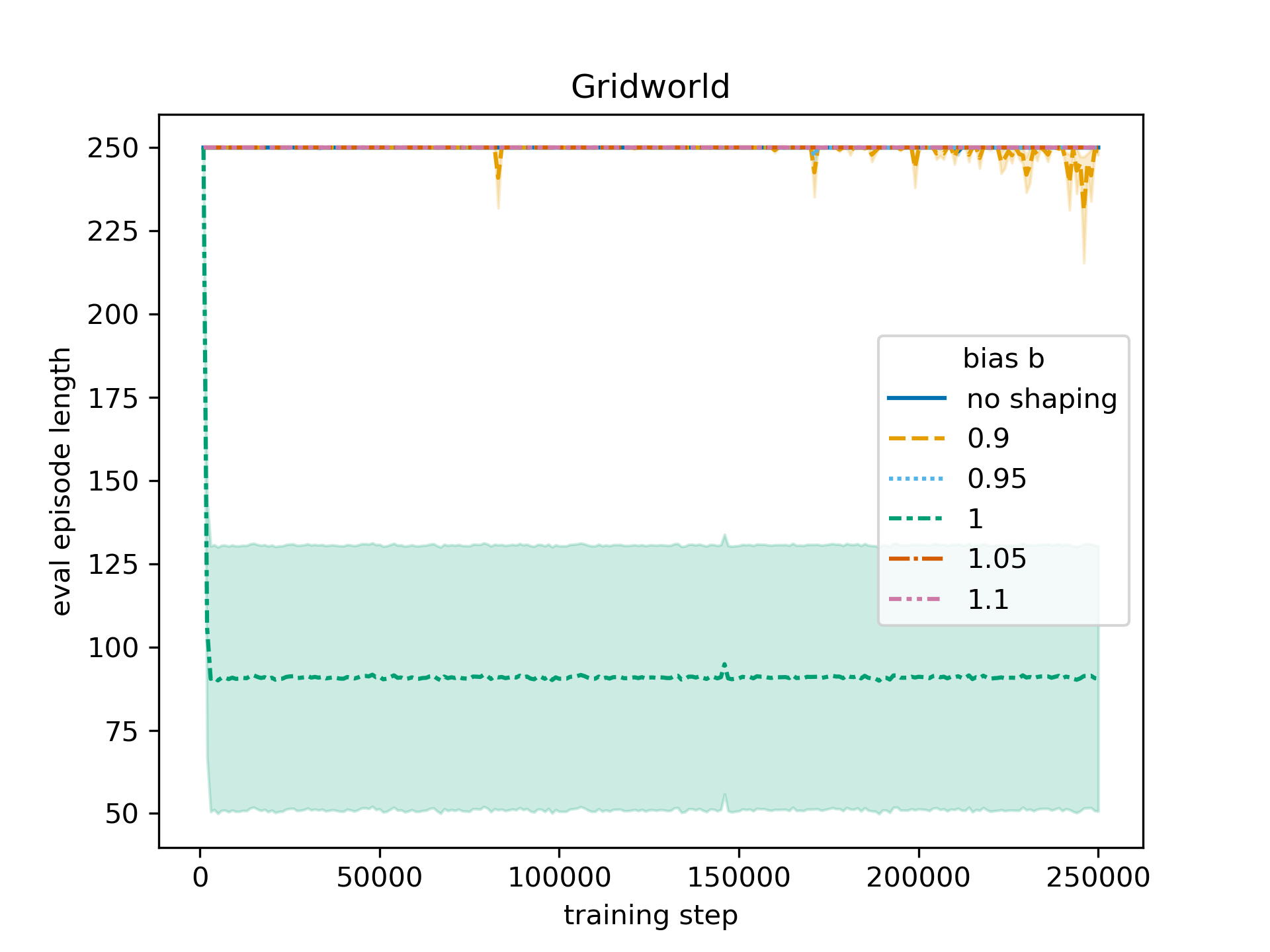}}
\subfigure[on-step, $Q_{init}$ = +1]{\includegraphics[width=0.33\textwidth]{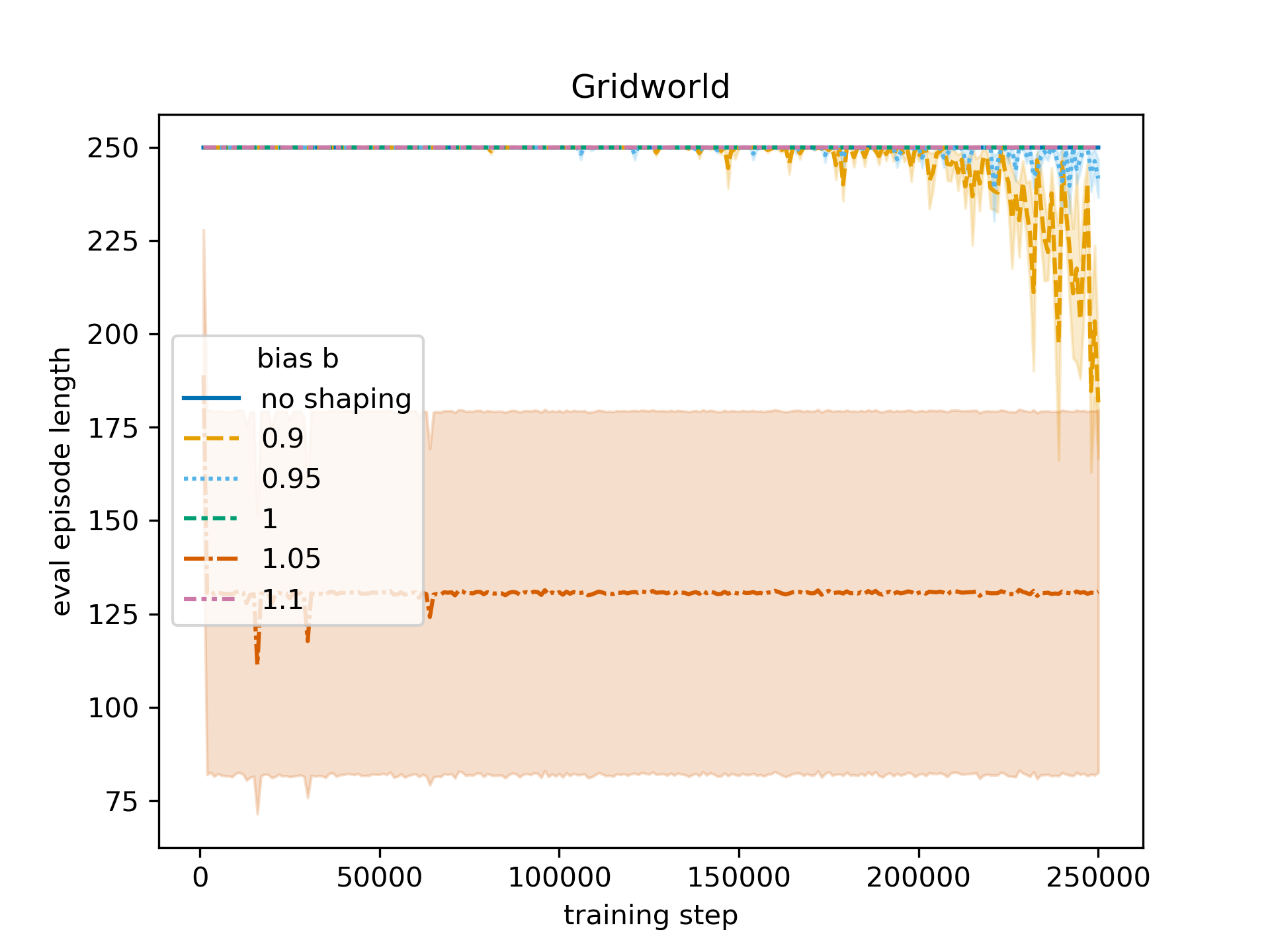}}
\caption{Gridworld results for the two different reward functions \textit{goal-directed} and \textit{on-step}. The figures in the top row show the results for the \textit{goal-directed} reward function. The figures in the bottom row show the results for the \textit{on-step} negative rewards. Each plot shows the length of the evaluation episodes for different values of the shifting bias $b$. Each graph showing the mean length of the ten evaluation runs with fixed exploration rate $\epsilon=0.05$ that were run every 250 training steps averaged over five separate training runs. Shaded areas showing the standard error of the mean.}
\label{fig:results-gridworld}
\Description[Gridworld results for the two different reward functions goal-directed and on-step plotting evaluation episode lengths over the training duration for different initial Q-values and values of the bias.]{Gridworld results for the two different reward functions goal-directed and on-step plotting evaluation episode lengths over the training duration for different initial Q-values and values of the bias. The goal-directed results show fast and reliable convergence only for the bias parameter that compensates for the respective initial Q-value. The on-step results also only converge for the bias value that compensates the initial Q-values, but show larger variance with some of the runs not converging at all.}
\end{figure*}

\subsubsection{Gridworld} In this environment the agent is tasked to navigate a discrete square grid. The agent starts in the top-left corner of the grid and has to move into the goal state in the bottom-right corner. Entering the goal state terminates the episode. Each episode has a maximum of 250 training steps. We experiment with two different reward functions for this environment that both lead to the same task and optimal policy. 

The first reward function named \textit{goal-directed} gives a reward of zero for every transition except when moving into the goal state where the agent gets a reward of one. 
The second reward function named \textit{on-step} matches the common rationale of offering a negative reward for each step to get the agent to minimize its steps. With it the agent gets a reward of minus one for each step. 

For this environment we use the inverse of the Manhattan distance to the goal as the potential function: $\Phi(s) = -d(s, s_g)$, which is then normalized to be in the usual range of $[0, 1]$. The potential-based reward shaping will therefore be able to directly leverage a perfect oracle of how to solve the task.

This first set of experiments is used to validate our theoretical results in section \ref{sec:shift} regarding the choice of the bias as $b = -R + (1-\gamma) Q_{init}$.
Figure \ref{fig:results-gridworld} shows the results for the experiments on the Gridworld environment for two different reward functions, \textit{goal-directed} where $R=0$ and \textit{on-step} where $R=-1$, for three different initial Q-values (-1, 0, +1) with a decay factor $\gamma=0.95$.

The results for the \textit{goal-directed} reward function are shown in the top row. The \textit{goal-directed} reward is zero for all transitions except when moving into the goal where the reward is one. The results show that for each combination of initial Q-values $Q_{init}$ and bias $b$ only the combination with a bias $b=(1-\gamma)Q_{init}$ converges to the optimal policy within the 250,000 training steps. Choosing the bias correctly leads to very quick convergence in this setting as the potential function is a perfect heuristic for solving the task. The agent therefore just has to choose the optimal action in each state randomly once before the respective Q-value is the largest one. With the usual advantage-based action selection schemes (like $\epsilon$-greedy) the agent will therefore choose the optimal action when it reaches the same state in future iterations and exploit the task knowledge properly.

If the choice of bias does not fit the initial Q-value the evaluation performance does not improve within the training steps. The problem for the convergence in these cases is the incorrect resulting positive or negative shaping rewards. In that case the agent will need to repeat all actions in all states multiple times to properly converge from the initial Q-values and the shaping signal can give incorrect intermediate incentives which leads to the exploitation of sub-optimal actions by the agent. 

The second set of results for the \textit{on-step} reward function is shown in the bottom row. Negative on-step rewards are commonly used to directly indicate to an agent to reach terminal states as early as possible even though the same requirement is also given indirectly by choosing a decay factor $\gamma < 1$. For our specific Gridworld environment the optimal policy is the same as when using the \textit{goal-directed} reward function as the goal state is the only state the agent can terminate in.

In this setting once again only one choice of bias can improve the sample efficiency and lead to converge towards the optimal policy within the limited number of training steps. But, the average performance over five separate repetitions show high variance in training performance for the optimal choices of bias. The large bias leads to large negative potential values which lead to large positive shaping rewards for any terminal state including when the episode is truncated after a fixed number of steps. As a result, some runs get stuck oscillating into the non-goal state the agent has truncated an episode in before. The runs in this setting therefore either converge very quickly or do not converge at all within the training budget. 

In theory over infinitely many training steps the agent would eventually execute sequences of random exploration actions that lead to the real goal-state often enough to be able to converge to the optimal policy. This however can only happen over a longer training duration. Longer episodes could also help in this case as they improve the likelihood of finding the goal state instead of the episode being truncated in a non-goal state.

Notably, the dense non-zero rewards can implicitly create an optimistic Q-value initialization, which leads to more systematic exploration and can therefore lead to faster convergence. As such, some of the runs with an incorrect bias choice can be seen to improve their evaluation performance towards the end of the training.

\begin{figure*}[htb]
\centering
\subfigure[Cart Pole]{\includegraphics[width=0.45\textwidth]{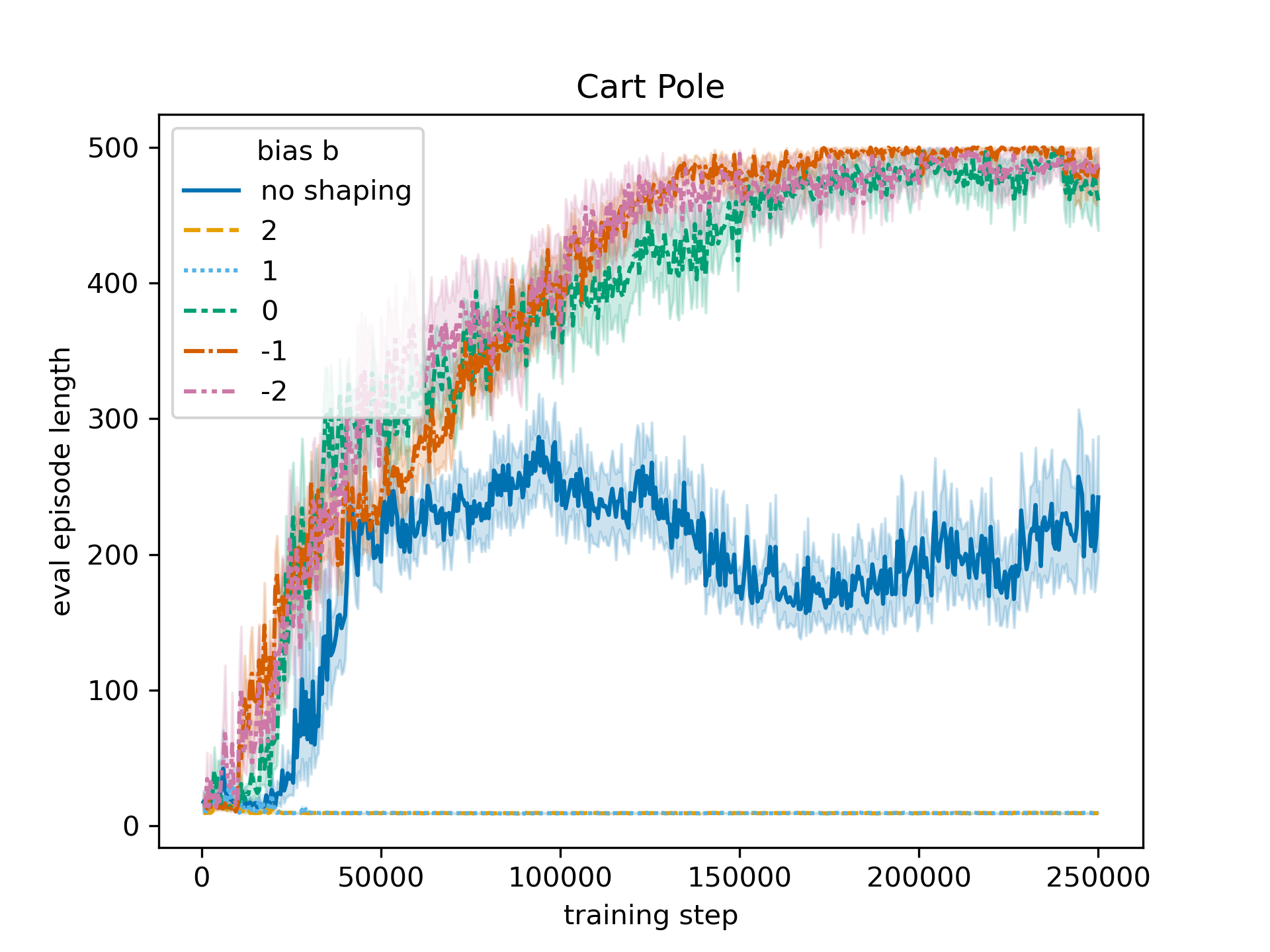}\label{fig:cart-pole-results}}
\subfigure[Mountain Car]{\includegraphics[width=0.45\textwidth]{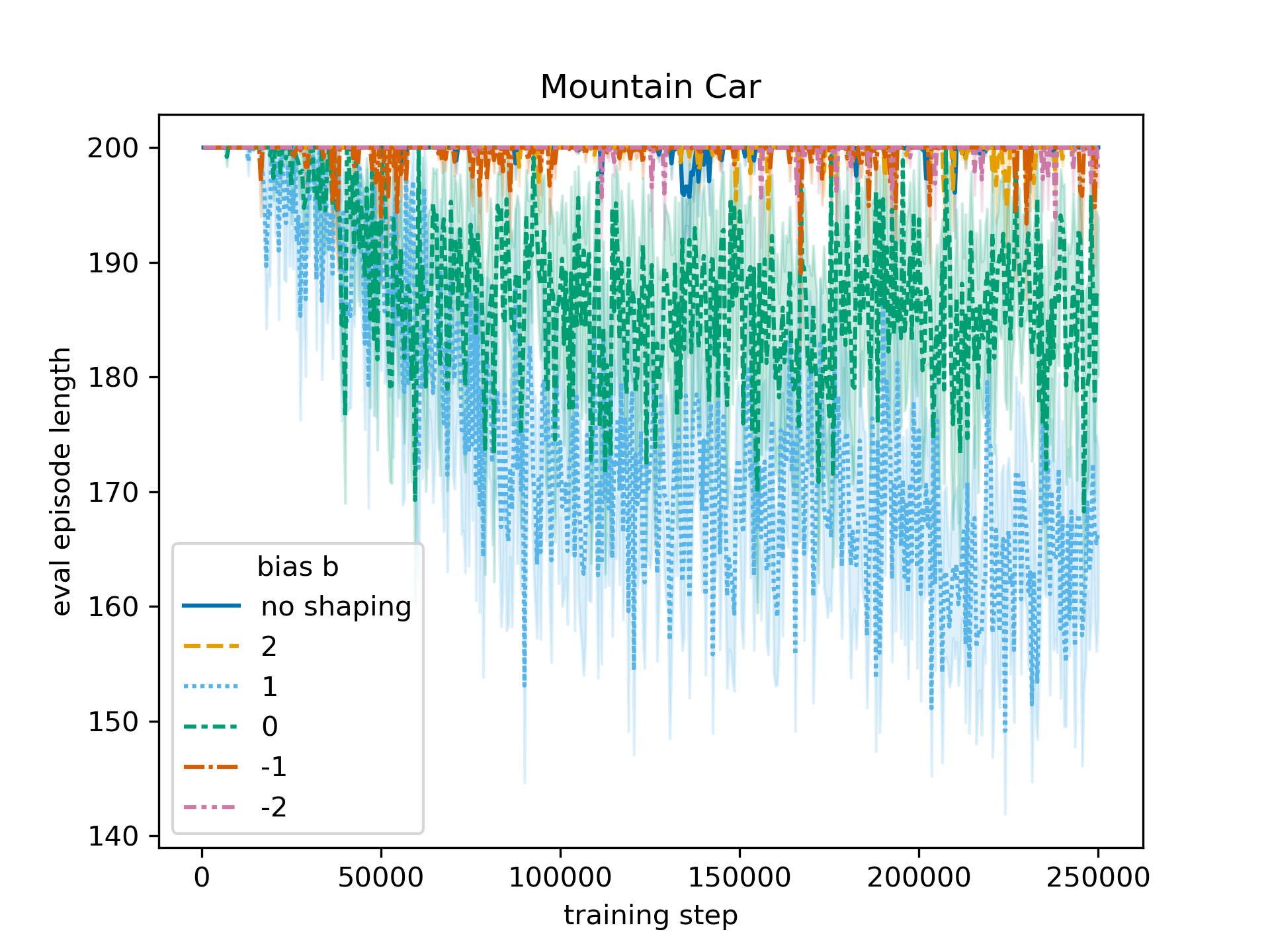}\label{fig:mountain-car-results}}
\caption{Results of Cart Pole and Mountain Car experiments for different values of the bias parameter $b$. Each plot showing the length of the evaluation episodes that were run every 500 steps averaged over five evaluation runs per training run and plotting the mean of ten separate training runs with the shaded area being the standard error of the mean.}
\label{fig:deep-results}
\Description[Results of Cart Pole and Mountain Car experiments for different values of the bias parameter.]{Results of Cart Pole and Mountain Car experiments for different values of the bias parameter plotting the average evaluation episode length over the training duration of 250,000 steps. In Cart Pole bias values greater or equals zero converge to the best policy. The not shaped baseline converges to a worse performing policy. Negative bias values lead to immediately terminating policies. In Mountain Car a bias of one leads to the best performing policy, a bias of zero to the second best performing policy. Other bias values and the not shaped baseline do not converge at all over the training duration.}
\end{figure*}

\subsubsection{Cart Pole}
In Cart Pole \citep{barto1983cartpole} the agent controls a cart with a pole attached to it that can slide across a track. The goal of the agent is to keep the pole upright (within a certain angle) and keep the cart within a predefined interval for as long as possible. At every time step the agent chooses which direction (left or right) to move the cart. Every time step gives a reward of one until one of the task conditions is violated, upon which the episode is terminated. Otherwise, the episode will be truncated after training 500 steps.

The potential function for this environment is the inverse of the observed pole angle: $ \Phi(s) = -|angle(s)|$, which is then normalized to be within $[0,1]$. Both the position on the track and the angle of the pole are part of the observation space of the agent. The potential function therefore encodes an important part of how to solve the task, but misses the additional limitation of staying within a fixed range on the track.

In this experiment, we test the theory in section \ref{sec:shift} for the choice of the bias parameter $b$ in a deep RL setting with a constant $+1$ reward.
Figure \ref{fig:cart-pole-results} plots the average evaluation performance for a selection of bias values in the Cart Pole environment.

When probing untrained DQNs with randomly sampled valid states for Cart Pole the initial Q-values were usually in $[-1, 1]$ with the exact distribution depending on the random network weight initialization. But, we do not focus on matching the bias values to possible initial Q-values as the Q-values for not visited states can change at any time during the training of the neural network. As such the Q-values for never before visited states might have already changed away from the initial Q-values.

The results show that the runs with a bias less than or equal to zero are able to leverage the potential function to reliably converge faster to a better policy than the agent without reward shaping. If the bias is chosen greater than zero the evaluation performance is consistently close to zero. As such our results for the potential shifting apply to the function approximation setting. Notably, in this simple environment the choice of the actual bias value does not matter much besides the general trend, as the agent is already able to learn well even without the additional information of reward shaping.

\subsubsection{Mountain Car}
In Mountain Car \citep{Moore1990mountaincar} the agent controls a car in a valley between two mountains with the goal of reaching the top of the right mountain. We use the environment definition with discrete actions. The agent can choose to either accelerate to the left or to the right, or not to accelerate at the current time step. As the task is to minimize the number of steps to reach the goal the agent is returned a reward of minus one for each step until the episode terminates. The episode is truncated after 200 training steps.

\balance
We use the absolute velocity of the cart as the potential function in this environment: $\Phi(s) = |velocity(s)|$, which is then normalized into $[0,1]$. The agent observes both its current position and its velocity at every time step. The potential function imperfectly guides the agent into states where the velocity is maximized, which is needed to be able to reach the goal. But the agent will have to learn a policy which decreases its velocity whenever the cart has to switch directions to build enough momentum to reach the goal.

In this experiment we test the theory in section \ref{sec:shift} for the choice of the bias parameter $b$ in a deep RL setting with a constant $-1$ reward.
Figure \ref{fig:mountain-car-results} shows the average evaluation performance of agents in the Mountain Car environment with differently shifted potential functions compared with an agent without additional reward shaping. For the same reason as in the Cart Pole experiment we do not test for the comparably small changes to the bias $b$ that a compensation of the initial Q-values would create.

In this environment only the two configurations with bias zero and one lead to a policy that is able to solve the task reliably. The bias of one leads to the reliable learning of the best performing policies. This matches our theoretical results in section \ref{sec:shift} of using the bias term to correct for the constant negative on-step rewards in this environment. When randomly sampling initial Q-values for valid states from untrained DQNs they tend to fall within [-1, 1]. Our results therefore also indicate that the Q-values do not change drastically in this environment. This environment is more susceptible to changes in bias than the Cart Pole environment, but matches the expected behavior regarding the improved performance when correcting for the constant non-zero on-step rewards.

Notably, the variance between the individual runs for the well working choices of the bias $b$ is small. The issue of intermediate convergence to sub-optimal terminal states seen for the optimal $b$ choices in the \textit{on-step} Gridworld do not appear to also apply to the deep RL setting.


\section{Conclusion}
We have introduced a framework of requirements for effective potential-based reward shaping and derived how the potential function must be structured to allow for effective reward shaping to guide exploration.  

We have shown that a constant shift of the potential function in potential-based reward shaping can help alleviate problems caused by a mismatch between the original reward and the initial Q-values to improve the effectiveness of reward shaping in guiding the agent, and thus the sample efficiency of an RL agent. We have determined how to choose the value for a bias for this shift depending on the known or observed values for the rewards for each step as well as the initial Q-values. We empirically verified that this approach also holds for function approximation in deep RL.

We have shown the theoretical limitations of potential-based reward shaping in terms of correctly assigning positive or negative shaping rewards for continuous potential functions when dealing with small changes in potential value between consecutive states.



\begin{acks}
This work was supported by the Lower Saxony Ministry of Science and Culture (MWK), in the zukunft.niedersachsen program of the Volkswagen Foundation (HybrInt).
\end{acks}



\bibliographystyle{ACM-Reference-Format} 
\bibliography{main}


\end{document}